\documentclass{article}
\usepackage{spconf,amsmath,graphicx}
\usepackage{times,amsmath,epsfig}
\usepackage{amsmath,amssymb,bm}
\usepackage{booktabs}
\def\x{{\mathbf x}}
\def\p{{\mathbf p}}
\def\e{{\mathbf e}}
\def\z{{\mathbf z}}
\def\s{{\mathbf s}}

\title{MAPPING EXTENDED LANDMARKS FOR RADAR SLAM}
\name{Shuai Sun$^{\star}$ \qquad Christopher Gilliam$^{\dagger}$ \qquad Kamran Ghorbani$^\ddagger$ \qquad Glenn Matthews$^\ddagger$ \qquad Beth Jelfs$^{\dagger}$}
\address{$^{\star}$ Navigation College,  Dalian Maritime University, Liaoning, China \\
$^{\dagger}$ Dept. Electronic, Electrical \& Systems Engineering, University of Birmingham, Birmingham, UK\\
$^{\ddagger}$ School of Engineering, RMIT University, Melbourne, Australia}

\begin{document}
% \ninept
%
\maketitle
\begin{abstract}
Simultaneous localization and mapping (SLAM) using automotive radar sensors can provide enhanced sensing capabilities for autonomous systems. In SLAM applications, with a greater requirement for the environment map, information on the extent of landmarks is vital for precise navigation and path planning. Although object extent estimation has been successfully applied in target tracking, its adaption to SLAM remains unaddressed due to the additional uncertainty of the sensor platform, bias in the odometer reading, as well as the measurement non-linearity. In this paper, we propose to incorporate the Bayesian random matrix approach to estimate the extent of landmarks in radar SLAM. We describe the details for implementation of landmark extent initialization, prediction and update. To validate the performance of our proposed approach we compare with the model-free ellipse fitting algorithm with results showing more consistent extent estimation. We also demonstrate that exploiting the landmark extent in the state update can improve localization accuracy.
\end{abstract}
%
%==================================================
\section{Introduction}
\label{sec:intro}
%==================================================
In recent years, there has been increased interest in simultaneous localization and mapping (SLAM) using automotive radar sensors, mainly due to its relatively high accuracy, low cost, and small size, as well as its robustness when operated in harsh weather conditions, such as heavy rain or dense fog~\cite{Sun2020MIMORadarSensor}. The 360 degree field of view (FoV) can be achieved by either using a combination of short, middle, and long range radar sensors (each with a different FoV)~\cite{Patole2017automotiveRadarProcessing}, or a 360 degree rotational radar sensor~\cite{VivetFMCW_SLAM_2013}. Radar SLAM has been applied in both indoor and outdoor environments. In particular, the success of employing automotive radar in vehicle-based applications such as assisted or autonomous driving demonstrates a promising research direction~\cite{Meinel2014,Waldschmidt2021,cheng2021USV_SLAM}.

One of the main methods for environment modeling in SLAM is landmark-based mapping, where sensed objects are abstracted as landmarks. The goal is thus to estimate the state of landmarks in order to construct a map of the region surrounding the mobile platform. In the early years, algorithms developed for landmark-based SLAM operated under the assumption that each landmark can generate at most one measurement per scan (point landmark), and therefore mainly focused on estimating the location of landmarks. However, with the advent of high resolution automotive radar sensors, landmarks may occupy multiple radar resolution cells and hence generating more than one radar detection per scan, i.e. an extended landmark. By exploiting information such as the location of each radar detection, as well as their spacial spread, we can also estimate the size and orientation of a landmark, in addition to its centroid location. 

Landmark extent estimation can better facilitate navigation and path planning for the mobile platform, since it provides a more precise map of the surrounding environment. However, existing methods of object extent estimation have mainly been developed in the target tracking literature (typically the single/multiple extended target(s) tracking~\cite{Beard2015multipleExtendedTargetTracking, Granstrom2022TutorialEoT}) and adaption of these approaches to radar SLAM is non-trivial. Firstly, the pose (location and orientation) of the mobile platform is unknown and evolves with time, and hence needs to be estimated simultaneously in SLAM. Compared with conventional extended target tracking using linear measurements and a fixed known sensor location, uncertainty in the estimated pose of the mobile platform and measurement conversion from polar coordinate to Cartesian coordinate give rise to additional complexity for SLAM in the landmark extent estimation. Secondly, due to the relative short detection range of an automotive radar, as well as the motion of the mobile platform (and potentially the motion of a landmark in a dynamic environment), a landmark may frequently move in and out of the radar FoV, preventing a consistent estimation in a timely manner. In addition, there are unique problems associated with automotive radars: radar detections are sparse at each scan, and are severely affected by the relative geometry (aspect angle) between the radar and the landmark. This makes recursive inference of landmark extent difficult, since we may lack sufficient measurement data.

In this paper, we propose a method for landmark extent estimation within the extended Kalman filter (EKF) SLAM framework, with the aim of enhancing mapping in radar SLAM applications. The landmark extent estimation is a further extension of our previous work~\cite{2022landmarkManagementRadarSLAM}, where only centroid location of each landmark is considered. Specifically, using our landmark management scheme, we introduce an extent prediction and update procedure for each registered landmark in the system. Accordingly, the main contributions of this paper are three-fold: 1) we implement the Bayesian random matrix algorithm (RMA) for landmark extent estimation in radar SLAM under the assumption of an elliptical model; 2) we demonstrate the effectiveness of the RMA for radar SLAM landmark extent estimation, and compare the results with the ellipse fitting algorithm for performance validation; 3) we apply the available landmark extent information in the association of new radar detections to existing landmarks, as well as better qualifying the measurement uncertainty due to the landmark extent.

%==================================================
\section{Problem Formulation}
\label{sec:formulation}
%==================================================
Typically, in the target tracking literature, landmark extent estimation methods differ in the following three aspects:
\begin{description}
    \item[Shape modeling.] By assuming some general parametric shape for the target extent, we can approximate different types of targets with varying sizes and shapes. Typical parametric shapes includes employing a specific geometric shape, such as en ellipse or a rectangle~\cite{Granstrom2022TutorialEoT}, and a more general star-convex model to represent an arbitrary shape~\cite{Lee2019StarConvexTSP}. 
    \item[Spread modeling.] Measurement spread refers to the spatial distribution of the generated radar measurements around the target extent, such as the uniform surface distribution model, the boundary (edge) distributed contour model~\cite{Lan2018ExtendedRectangularRadar}, the random hypersurface model~\cite{Baum2014ExtendedHypersurfaceModel}, and the reflection/scattering point-based model~\cite{Lars2012ReflectorCenterModel}. In practice, measurement spread distribution depends on many factors, including the sensor type, target structure and material, as well as the aspect angle between the sensor and the target. 
    \item[Algorithm.] Algorithms developed for target extent estimation vary with their assumptions on target shapes and the spatial spread in measurement data. Well-known algorithms include the Bayesian random matrix algorithm~\cite{Feldmann2010RadomMatrixTracking}, extended Kalman filter based algorithm~\cite{Yang2019SurfaceModel_multplicativeNoise}, and the random finite set based algorithm~\cite{Karl2014ExtendRFS}.
\end{description}
In this work we employ an ellipse shape for landmark extent modelling, assume a uniform surface distribution model and use the Bayesian random matrix algorithm to estimate the landmark extent. In the following, we outline our approach and refer readers to our previous work~\cite{2022landmarkManagementRadarSLAM} for details on the landmark management scheme under the EKF-SLAM framework.

\begin{figure}[tb]
	\centering
	\centerline{\includegraphics[width=.92\linewidth]{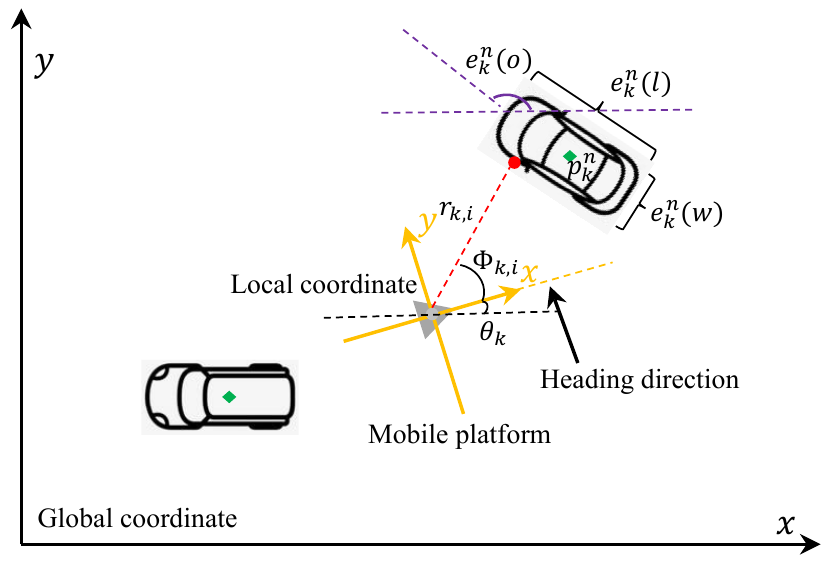}}
	\caption{Illustration of a typical considered SLAM scene.}
	\label{fig:scheme}
\end{figure}

Figure~\ref{fig:scheme} depicts a typical setup for landmark-based radar SLAM. At each time $k$, we aim to jointly estimate the pose of the mobile platform, namely the 2D location and heading of the mobile platform, $\x_k^m=\left[x_k,\ y_k,\ \theta_k\right]^T$ and the information of each sensed landmark $n$, namely the centroid location, $\p_k^n = \left[p_k^n(x),\ p_k^n(y)\right]$ and the physical extent $\e_k^n = [e_k^n(l),\ e_k^n(w),\ e_k^n(o)]$ representing the length, width and orientation respectively. 

Following the formulation of the EKF-SLAM, we define an augmented state vector for time $k$, denoted by $\x^a_k$, as
\begin{equation}
 \x_k^a = \left[\x_k^m,\ \x_k^{\ell}\right]^T,
\end{equation}
where $\x_k^{\ell} = \left[\p^1_k, \dots, \p_k^{N_k}\right]^T$ is the vector of landmarks registered in the system. As well as $\x_k^{\ell}$, we additionally define a corresponding landmark extent vector $\e_k^{\ell}$ to encompass the extent of all of the landmarks, 
\begin{equation}
	\e_k^{\ell} = \left[\e_k^1,\ \e_k^2, \cdots, \e_k^{N_k}\right]^T.
\end{equation}
Note the total number of landmarks $N_k$ will vary with time as newly confirmed landmarks will be added and false landmarks will be removed.

%--------------------------------------------------
\subsection{Measurement Model}
%--------------------------------------------------
We assume there are $m_k^n$ radar detections obtained from landmark $n$ at time $k$, where each radar sensor measurement $\s_{k,i}^n= \left[r_{k,i}^n,\ \phi_{k,i}^n\right]$ consists of the range, $r_{k,i}^n$, and azimuth, $\phi_{k,i}^n$, relative to the mobile platform, where $i=1,\ldots,m_k^n$. For object extent (shape and size) estimation, a conversion of the sensor measurements from the polar coordinate to a 2D Cartesian coordinate is usually required, giving
\begin{align}
    \label{eq-xy_measurement}
    \z_{k,i}^{n} = f\Big(\x_k^m,\ \s_{k,i}^n\Big) = \begin{bmatrix}
		x_k + r_{k,i}^n \cos\left(\theta_k + \phi_{k,i}^n\right)  \\ 	
		y_k + r_{k,i}^n \sin\left(\theta_k + \phi_{k,i}^n\right) 
	\end{bmatrix}.
\end{align}
The joint distribution of the converted Cartesian measurements $\z_k^n$, generated from landmark $n$, is modelled as
\begin{equation}
	p\Big(\z_k^n \Big|\ \x_k^a,\ \s_{k,i}^n\Big) = \prod_{i=1}^{m_k^n} \mathcal{N} \left(\z_{k,i}^{n};\ \p_k^n,\ \gamma_z X_k^n + W_k^n\right).
\end{equation}
Here $\gamma_z$ is a scaling factor accounting for the spread contribution of the landmark extent $X_k^n$, where $X_k^n$ is a random matrix modeling the landmark extent (see~\cite{Feldmann2010RadomMatrixTracking} for a detailed explanation). $W_k^n$ is the covariance of Cartesian measurements which originates from both the uncertainty in the platform pose (denoted by $P_k^m$ and obtained from the covariance of $\x_k^m$) and the sensor noise covariance $R_k$. The non-linearity of $f\left(\x_k^m,\ \s_{k,i}^n\right)$ with respect to the pose of the mobile platform and the original radar measurements gives rise to difficulties in calibrating $W_k^n$. For each measurement, $W_{k}^n(i)$ can be approximated as
\begin{equation}
    W_{k}^n(i) = \nabla f_{\x_k^m} P_k^m \nabla f_{\x_k^m}^T + \nabla f_{\s_{k,i}^{n}} R_k \nabla f_{\s_{k,i}^{n}}^T,
\end{equation}
where $\nabla f_{\x_k^m}$ and $\nabla f_{\s_{k,i}^{n}}$ are the Jacobians of $f\left(\x_k^m,\ \s_{k,i}^{n}\right)$ with respect to $\x_k^m$ and $\s_{k,i}^{n}$, as shown below,
\begin{align}
	\nabla f_{\x^m_k} & = \left.\dfrac{\partial f}{\partial \x^m_k}\right|_{(\x_{k}^m, \s_{k,i}^{n})} = 
	\begin{bmatrix}
	    1 & 0 & -r_{k,i}^n \sin(\theta_k + \phi_{k,i}^n)  \\ 
	    0 & 1 &  r_{k,i}^n \cos(\theta_k + \phi_{k,i}^n) 
	\end{bmatrix}.\\
	\nabla f_{\s_{k,i}^{n}} &= \left.\dfrac{\partial f}{\partial \s_{k,i}^{n}}\right|_{(\x_{k}^m, \s_{k,i}^{n})} \nonumber\\
	&= \begin{bmatrix}
	    \cos(\theta_k + \phi_{k,i}^n) & -r_{k,i}^n \sin(\theta_k + \phi_{k,i}^n)  \\ 
	    \sin(\theta_k + \phi_{k,i}^n) & r_{k,i}^n \cos(\theta_k + \phi_{k,i}^n)
	\end{bmatrix}.
\end{align}
We thus approximate $W_{k}^n$ as the maximum of $\Big\{W_k^n(i)\Big\}_{i=1}^{m_k^n}$.

%==================================================
\section{Landmark extent estimation}
\label{sec:algorithm}
%==================================================
To model the physical extent of landmark $n$ as an ellipsoidal shape at time $k$, we use a semi-positive definite (SPD) random matrix $X_k^n$ and adapt the Bayesian random matrix solution in~\cite{Feldmann2010RadomMatrixTracking} to estimate the landmark extent. Conversion between $X_k^n$ and its extent parameter $\e_k^n$ can be found in~\cite{Li2014EllipseFitting} (Equations (8-9)). In the following, we provide a summary of the steps for implementation of the landmark extent estimation.

%--------------------------------------------------
\subsection{Initialization}
%--------------------------------------------------
To initialize $\e_k^n$ for each registered landmark we apply the ellipsoid fitting based approach (EFA)~\cite{Li2014EllipseFitting}, using the converted measurements from~\eqref{eq-xy_measurement}. Once a new landmark is registered (only the centroid location $\p_k^n$ is available), its extent initialization may be postponed until the number of accumulated measurements from recent associated radar detections exceeds a pre-defined threshold $N_i$. Then the landmark orientation $e_k^n(o)$ and semi-axis length $[e_k^n(l),\ e_k^n(w)]$ can be estimated respectively. To reduce computation, we use the upper bounds in~\cite{Li2014EllipseFitting} as the estimated semi-axis length, since the initialization is a coarse step for landmark extent estimation. 

%--------------------------------------------------
\subsection{Extent Prediction}
%--------------------------------------------------
For the purposes of the extent prediction we assume the landmark extent remains constant over time, 
\begin{equation}
    X_{k|k-1}^n = X_{k-1|k-1}^n.
\end{equation}
As for the variance of the extent estimation, we follow the assumption that it may increase over time based on the choice of a temporal decay constant $\tau$,
\begin{equation}
    \alpha_{k|k-1}^n = 2 + \exp\Big(-\Delta t/\tau\Big)\Big(\alpha_{k-1|k-1}^n - 2\Big),
\end{equation}
where $\Delta t$ is the prediction time interval.

%--------------------------------------------------
\subsection{Extent Update}
%--------------------------------------------------
The mean of the converted Cartesian measurement  $\bar{\z}_k^n$ and its spatial spread $\bar{Z}_k^n$ are computed as
\begin{align}
    \bar{\z}_k^n & = \frac{1}{m_k^n} \sum_{i=1}^{m_k^n} \z_{k,i}^n,  \\
    \bar{Z}_k^n & = \sum_{i=1}^{m_k^n} \left(\z_{k,i}^n - \bar{\z}_k^n\right)\cdot\left(\z_{k,i}^n - \bar{\z}_k^n\right)^T.
\end{align}
The predicted variance of a single measurement $Y_{k|k-1}^n$ and the covariance of the mean measurement $M_{k|k-1}^n$ are
\begin{equation}
    Y_{k|k-1}^n = \gamma_z X_{k|k-1}^n + W_k^n,
\end{equation}
\begin{equation}
    M_{k|k-1}^n = (\bar{\z}_k^n - \p_k^n) (\bar{\z}_k^n - \p_k^n)^T.
\end{equation}
The innovation covariance $S_{k|k-1}^n$ is approximated as
\begin{equation}
    S_{k|k-1}^n = P_{k|k-1}^n + \frac{Y_{k|k-1}^n}{m_k^n}.
\end{equation}
If we define 
\begin{align}
    A_k^n &= (X_{k|k-1}^n)^{\frac{1}{2}} (S_{k|k-1}^n)^{-\frac{1}{2}}, \\
    B_k^n &= (X_{k|k-1}^n)^{\frac{1}{2}} (Y_{k|k-1}^n)^{-\frac{1}{2}},
\end{align}
then the estimated variance and covariance become
\begin{align}
    \hat{M}_{k|k-1}^n &= A_k^n M_{k|k-1}^n (A_k^n)^T,\\
    \hat{Y}_{k|k-1}^n &= B_k^n \bar{Z}_k^n (B_k^n)^T.
\end{align}
Finally, the updated extent estimate for landmark $n$ is
\begin{equation}
    X_{k|k}^n = \frac{1}{\alpha_{k|k}^n} \left(\alpha_{k|k-1}^n X_{k|k-1}^n +  \hat{M}_{k|k-1}^n + \hat{Y}_{k|k-1}^n\right),
\end{equation}
where $\alpha_{k|k}^n$ is updated as
\begin{equation}
    \alpha_{k|k}^n = \alpha_{k|k-1}^n + m_k^n.
\end{equation}

%--------------------------------------------------
\subsection{Exploitation of Landmark Extent}
\label{ssec:Exploit}
%--------------------------------------------------
The estimated extent of the landmarks can be used to assist the EKF-SLAM in two aspects. First, at each time step new radar measurements need to be associated with existing landmarks, which is achieved via a sifting step, as described in~\cite{2022landmarkManagementRadarSLAM}. Having the estimated landmark extent allows the sifting to be conducted by checking whether a measurement lies within the contour of the landmark, without needing to resort to a sifting threshold. Secondly, the estimated landmark extent can be used to give an indication of the measurement spread associated with a landmark. This spread can then be used to calibrate the measurement uncertainty arising from the landmark extent during the EKF update step.

%==================================================
\section{Simulation results}
\label{sec:simulation}
%==================================================
To assess the performance of our proposed landmark extent approach under the EKF-SLAM framework, we use the same simulated car park environment as in~\cite{2022landmarkManagementRadarSLAM} (note that the same model parameters are used if not specified). We set the parameters used in RMA as $N_i=20$, $\tau=100$, $\alpha_0=50$, $\gamma_z = \frac{1}{4}$ and use the model-free EFA as a baseline comparison. Once a landmark extent is initialized, both approaches are applied at each time step for landmark extent estimation. To assess the performance we use a Monte Carlo simulation with $1000$ independent trials. The Gaussian Wasserstein distance (GWD)~\cite{Yang2016EllipseDistance} is used to quantify the difference between the ground truth landmark extent and the estimated extent.

\begin{figure}[tb]
	\centering
	\centerline{\includegraphics[width=\linewidth]{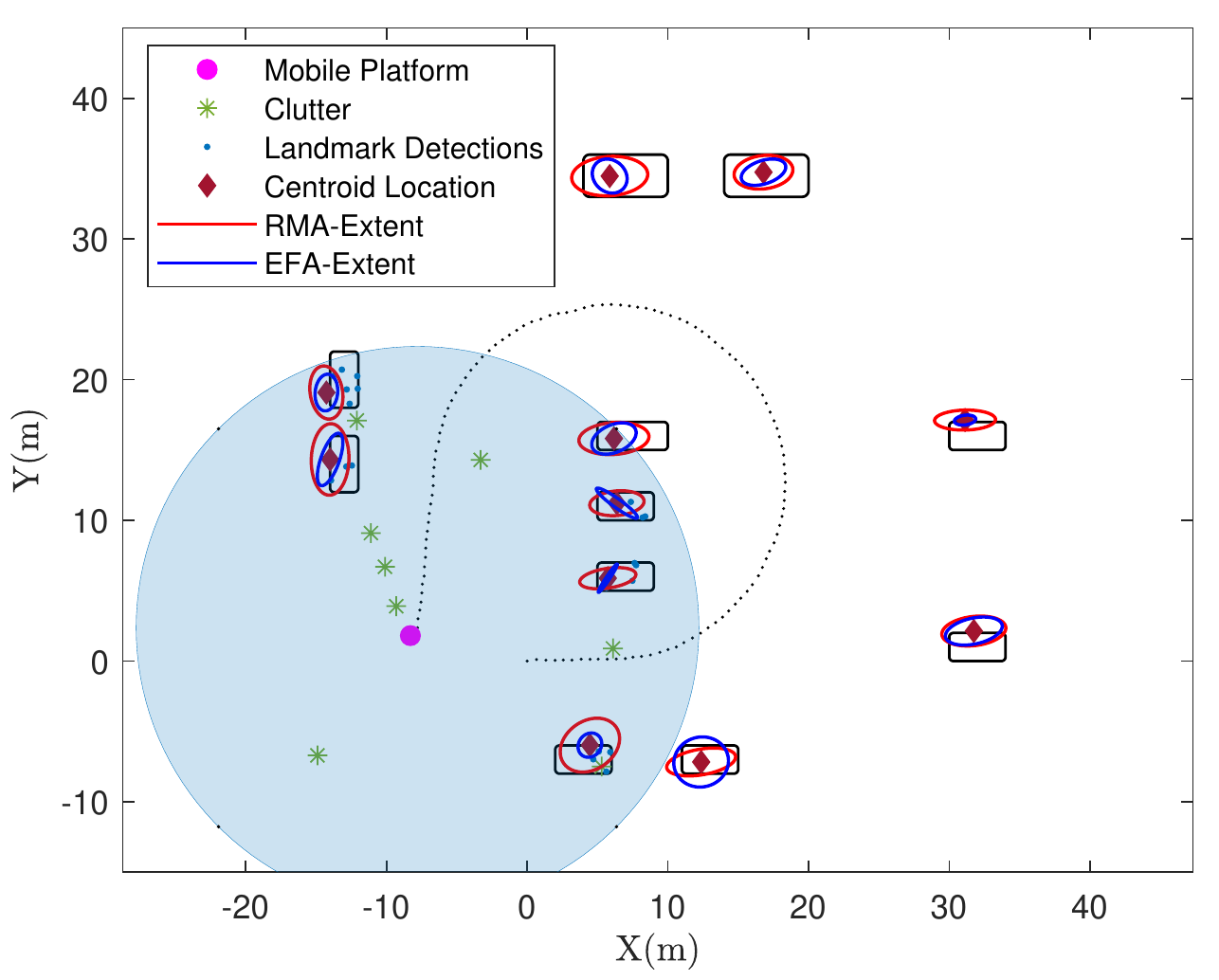}}
	\caption{Example of the performance for the RMA based (red ellipse) and EFA based (blue ellipse) solutions.}
	\label{fig:result}
\end{figure}

Figure~\ref{fig:result} shows a typical example of the end of a simulation, illustrating that both the RMA and EFA can be used to estimate landmark extent in EKF-SLAM. The two algorithms use the same initialization approach: once a landmark is registered in the system with a centroid estimate, its corresponding extent will be initialized when the accumulated number of associated measurements exceeds $N_i$.

\begin{figure}[tb]
    \centering
	\centerline{\includegraphics[width=\linewidth]{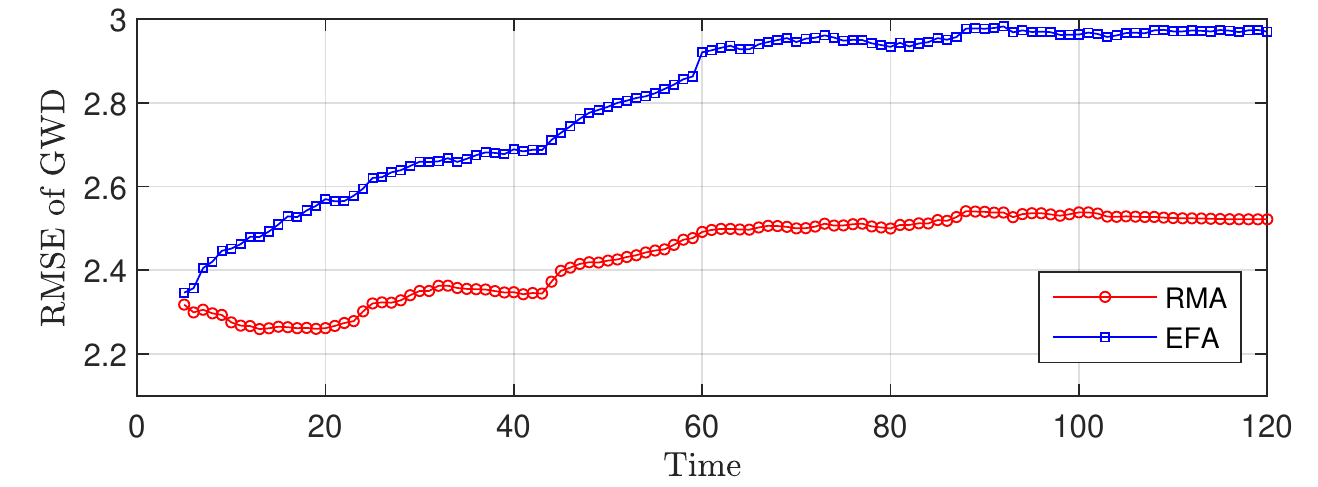}}
	\caption{Comparison of results based on the GWD metric.}\vspace*{-1.2mm}
	\label{fig:distance}
\end{figure}

The performance of the algorithms was compared using the root mean square error (RMSE) of the average GWD, with the statistical results shown in Fig.~\ref{fig:distance}. It can be seen from Fig.~\ref{fig:distance}, that overall the RMA maintains a more consistent extent estimation performance, compared with the EFA. The RMA employs a Bayesian scheme to exploit the previous estimate when updating the landmark extent estimate. The EFA, however, is a pure data fitting approach.

\begin{table}[tb]
    \caption{Platform pose error when incorporating landmark extent in the EKF-SLAM framework.} 
    \centering 
    \begin{tabular}{lcc} 
        \toprule 
        \textbf{Metric} & \textbf{No Extent} & \textbf{With Extent} \\
        \midrule
        Position avg. RMSE (m) & 1.21 & 1.09 \\
        Heading avg. RMSE ($^\circ$) & 3.48  & 3.24 \\
        % Landmark centroid avg. error (m) & 1.18 & 1.20 \\
        \bottomrule
    \end{tabular}
    \label{table:MC} 
\end{table}

In order to illustrate the benefit of the landmark extent on the EKF-SLAM, we compare the performance without using the estimated landmark extent to that achieved when exploiting the extent as described in Section~\ref{ssec:Exploit}. Table~\ref{table:MC} shows there is an improvement in the average error in the pose of the mobile platform, both in terms of position and heading. Note that although the improvement in this case is only small, one of the key benefits of exploiting landmark extent lies in the fact that it can help to reduce the dependence on empirical thresholds, which can be difficult to set in practice. 

%==================================================
\section{Conclusions}
\label{sec:conclusion}
%==================================================
In this paper, we propose a random matrix approach for the estimation of landmark extent for radar SLAM. For each registered landmark in the system we detail the extent initialization, prediction and update steps. Using a simulated car park environment we demonstrate the proposed method is able to maintain a consistent landmark extent estimation. Furthermore, we show that including the estimated landmark extent in the measurement association and EKF update steps can improve the localization of the mobile platform. These are initial results indicating that there is a benefit to exploiting the landmark extent when performing self localization and sensing tasks.

\vfill\pagebreak

\bibliographystyle{IEEEbib}
\bibliography{refs}

\end{document}